# Bacterial foraging optimization based brain magnetic resonance image segmentation

Abdul Kayom Md Khairuzzaman[1, *]

[1]Department of Electrical Engineering, National Institute of Technology, Silchar, India
[*]kayomabdul@gmail.com

**Abstract:** Segmentation partitions an image into its constituent parts. It is essentially the pre-processing stage of image analysis and computer vision. In this work, T1 and T2 weighted brain magnetic resonance images are segmented using multilevel thresholding and bacterial foraging optimization (BFO) algorithm. The thresholds are obtained by maximizing the between class variance (multilevel Otsu method) of the image. The BFO algorithm is used to optimize the threshold searching process. The edges are then obtained from the thresholded image by comparing the intensity of each pixel with its eight connected neighbourhood. Post processing is performed to remove spurious responses in the segmented image. The proposed segmentation technique is evaluated using edge detector evaluation parameters such as *figure of merit*, *Rand Index* and *variation of information*. The proposed brain MR image segmentation technique outperforms the traditional edge detectors such as canny and sobel.

**1. Introduction**

Segmentation partitions an image into certain regions of interest or clusters. It has numerous applications ranging from computer vision to target matching for military applications. Another interesting application is in medical science. For example, image segmentation helps in diagnosing abnormality in brain or any part of the body from the MRI or PET scan. MRI is a powerful non-invasive technique for diagnosis and treatment planning of various diseases such as multiple sclerosis (MS), Alzheimer's disease, Parkinson's disease, epilepsy, cerebral atrophy, presence of any lesion like glioma etc [1]. Effective image segmentation helps in classifying and analysing these disorders.

Histogram based thresholding is the most popular and simple technique for image segmentation. Otsu [2] proposed a method for automatic threshold selection by maximizing the between class variance in a gray level image. Kapur's [3] method utilizes maximization of posterior entropy that indicates homogeneity of the segmented classes. In general, the Kapur and Otsu methods are known for their better shape and uniformity measures. These methods are originally developed for bi-level thresholding and later on extended to multilevel thresholding. All these methods have one common problem that the computational complexity increases exponentially when extended to multilevel thresholding. This is due to the exhaustive search for the optimal thresholds, which limits their usage for multilevel thresholding applications [4].

Nature has always been an inspiration in solving computationally complex problems. Particle swarm optimization (PSO) [5, 6], Ant colony optimization (ACO) [7] and BFO [8] algorithms are inspired from the foraging behaviour of natural animals. Computational complexity of multilevel thresholding is greatly reduced by using these nature inspired optimization algorithms.

Genetic Algorithms (GA) [9], Particle Swarm Optimization (PSO) have been successfully applied in multilevel thresholding [10, 11]. Sathya et al. [4] showed that BFO algorithm performs better than PSO and GA for multilevel thresholding in terms of accuracy, speed and stability of the solution.

Bacteria Foraging Algorithm proposed by Passino [8], mimicked the foraging behaviour of E. Coli bacteria present in human intestine. In foraging theory, it is assumed that the objective of the animal is to search for and obtain nutrients such that the energy intake per unit time is maximized [8]. Maitra et al. [1] applied BFO algorithm with Kapur method for multilevel thresholding of brain MR Image and showed its superiority over PSO based multilevel thresholding.

Bi-level thresholding assumes that an image contains an object and the background. It is too harsh on an image and eliminates lot of information and hence fails in most of the cases. In case of medical image segmentation one has to be very careful while eliminating any information as it may affect diagnosis. Therefore, it is necessary to perform multilevel thresholding which retains most of the important information to help better diagnosis. However, multilevel thresholding without optimization is time consuming. MRI is state of the art technique for diagnosis of various diseases. An effective segmentation of MR image is therefore necessary for better diagnosis.

Our main objective is to determine optimal thresholds, so that the image can be subdivided into several classes with different gray levels for their easier analysis and interpretation. For this, we propose a multilevel thresholding technique based on multilevel Otsu method optimized with BFO algorithm for brain MR Image segmentation. First Multilevel thresholding is performed on the brain MR image. The edges in the thresholded image are then detected by comparing the intensity of each pixel with its eight connected neighbourhood. The proposed algorithm is then evaluated objectively using *Rand index* [12], *variation of information* and *Pratt Figure of Merit* [13]. For objective evaluation, a reference image is created by manually segmenting the original image. The proposed brain MR image segmentation technique out performs the traditional edge detection techniques such as Canny and Sobel.

## 2. Multilevel thresholding

Thresholding is a popular image segmentation technique because of its simplicity and effectiveness. Normally, image histogram is used to determine the thresholds. There exist a number of methods for threshold selection. Kapur et al. [3] and Otsu [2] are the most widely acknowledged histogram based automatic threshold selection methods. Kapur's method is based on maximizing posterior entropy of the thresholded image whereas Otsu method maximizes the variance between segmented classes. Here we explain briefly the Otsu method extended for multilevel thresholding [4, 2].

Let there be $L$ gray levels in a given image, i.e. $\{0,1,2,...,(L-1)\}$ and let us define $P_i = \frac{h(i)}{N}$, $(0 \leq i \leq (L-1))$ where $h(i)$ = number of pixels with gray level $i$ and $N$ = total number of pixels in the image $= \sum_{i=0}^{L-1} h(i)$.

Now, the multilevel thresholding problem can be configured as an m-dimensional optimization problem for determination of $m$ optimal thresholds $[t_1, t_2, ..., t_m]$ which divide the original image into $m+1$ classes: $C_0$ for $[0,...,t_1-1]$, $C_1$ for $[t_1,...,t_2-1]$,... and $C_m$ for $[t_m,...,L-1]$. The thresholds are obtained by maximizing the following objective function:

$$J(t_1, t_2, ..., t_m) = \sigma_0^2 + \sigma_1^2 + \sigma_2^2 + ... + \sigma_m^2 \quad \ldots\ldots\ldots (1)$$

Where
$\sigma_0^2 = \omega_0 (\mu_0 - \mu_T)^2$,
$\sigma_1^2 = \omega_1 (\mu_1 - \mu_T)^2$,
$\sigma_2^2 = \omega_2 (\mu_2 - \mu_T)^2$,...,
$\sigma_m^2 = \omega_m (\mu_m - \mu_T)^2$ are the variances of the segmented classes. Where the class probabilities are,

$$\omega_0 = \sum_{i=0}^{t_1-1} P_i, \quad \omega_1 = \sum_{i=t_1}^{t_2-1} P_i, \quad \omega_2 = \sum_{i=t_2}^{t_3-1} P_i, ..., \omega_m = \sum_{i=t_m}^{L-1} P_i$$

And the mean levels $\mu_0$, $\mu_1$, ..., $\mu_m$ for classes $C_0$, $C_1$, ..., $C_m$ are as follows:

$$\mu_0 = \sum_{i=0}^{t_1-1} \frac{ip_i}{\omega_0}, \quad \mu_1 = \sum_{i=t_1}^{t_2-1} \frac{ip_i}{\omega_1}, ..., \mu_m = \sum_{i=t_m}^{L-1} \frac{ip_i}{\omega_m}$$

Let $\mu_T$ be the mean intensity for the whole image, then we have,

$\omega_0\mu_0 + \omega_1\mu_1 + \omega_2\mu_2 + ... + \omega_m\mu_m = \mu_T$ And $\omega_0 + \omega_1 + \omega_2 + ... + \omega_m = 1$.

## 3. Bacterial Foraging Optimization

Foraging strategies are methods of locating, handling and ingesting food. Natural selection eliminates animals with poor foraging strategies. This facilitates the propagation of genes of most successful foraging strategies. After many generations, the poor foraging strategies are either eliminated or redesigned into better ones. A foraging animal looks to maximize energy intake per unit time spent on foraging within its environmental and physiological constraints.

The E. Coli bacteria present in human intestine follows foraging behaviour, which consists of *chemotaxis*, *swarming*, *reproduction* and *elimination or dispersal*. Passino [8] has modelled this evolutionary technique as an effective optimization tool.

*3.1 Chemotaxis*

The bacterial movement of swimming (in a predefined direction) and tumbling (altogether in different directions) in presence of attractant and repellent chemicals from other bacteria is called chemotaxis. A chemotactic step is a tumble followed by a tumble or run. To represent a tumble, a unit length random direction, $\varphi(j)$ is generated which is then used to model chemotaxis as follows,

$$X^i(j+1,k,l) = X^i(j,k,l) + C(i)\varphi(j) \quad \text{.......... (2)}$$

Where $X^i(j,k,l)$ represents the $i^{th}$ bacterium at $j^{th}$ chemotactic, $k^{th}$ reproductive and $l^{th}$ elimination or dispersal event. $C(i)$ is the step size in the direction of movement specified by tumble (run length unit).

*3.2 Swarming*

Bacterium which reaches a good food source produces chemical attractant to invite other bacteria to swarm together. While swarming, they maintain a minimum distance between any two bacteria by secreting chemical repellent. Swarming is represented mathematically as,

$$Jcc(X, P(j,k,l)) = \sum_{i=1}^{S} J^{i}cc(X, X^{i}(j,k,l))$$

$$= \sum_{i=1}^{S} [-d_{attract} \exp(-\omega_{attract} \sum_{n=1}^{m} (X_n - X^i)^2)]$$

$$+ \sum_{i=1}^{S} [h_{repellant} \exp(-\omega_{repellant} \sum_{n=1}^{m} (X_n - X^i)^2)] \quad \ldots\ldots (3)$$

Where $Jcc(X, P(j,k,l))$ is the value of the cost function to be added to the actual cost function to be optimized to simulate swarming behaviour. $S$ is the total number of bacteria, $m$ is the number of parameters (dimension of the optimization problem) to be optimized. $d_{attract}$, $\omega_{attract}$, $\omega_{repellant}$, $h_{repellant}$ are the coefficients to be chosen properly.

*3.3 Reproduction*

After completion of $N_c$ chemotactic steps, a reproductive step follows. Health of $i^{th}$ bacterium is determined as,

$$J^{i}_{health} = \sum_{j=1}^{Nc} J_{sw}(i,j,k,l) \quad \ldots\ldots (4)$$

Then, the bacteria are sorted in descending order of their health. The least healthy bacteria die and the other healthier bacteria take part in reproduction. In reproduction, each healthy bacterium splits into two bacteria each containing identical parameters that of the parent keeping population of bacteria constant.

*3.4 Elimination and dispersal*

The bacterial population in a habitat may change gradually due to constraint of food or suddenly, due to environmental or any other factor. All the bacteria in a region may be killed or a group may be dispersed into a new location. It may have the possibility of destroying chemotactic progress, but it also has the possibility of assisting chemotaxis, since dispersal event may place the bacteria to near good food sources [1].

*3.5 Bacterial foraging optimization algorithm*

The original BFO algorithm given by Passino [8] is modified by Sathya et al. [4] which increases the convergence speed and global searching ability of the algorithm. In the modified BFO algorithm, they take the global best bacterium for the calculation of swarm attraction function. Instead of averaging all the objective function values, they consider the best value for each bacterium. Let us briefly explain the BFO algorithm in step by step manner.

**Step 1**

Initialize the number of variables to be optimized $p$, number of E. Coli bacteria $S$, number of chemotactic steps $N_c$, maximum swimming length $N_s$, number of reproduction steps $N_{re}$, number of elimination or dispersal events $N_{ed}$, the probability of elimination or dispersal $P_{ed}$, and the step size $C(i)$, $i = 1, 2, ...S$.

**Step 2**

Elimination-dispersal loop: $ell = ell + 1$

**Step 3**

Reproduction loop: $k = k + 1$

**Step 4**

Chemo-taxis loop: $j = j + 1$

**Step 4.1**

For $i = 1, 2, ..., S$ take a chemotactic step for $i^{th}$ bacterium as:

**Step 4.2**

Calculate the value of objective function $J(i, j, k, ell)$

**Step 4.3**

Find the global best bacterium $X_{gn}$, from all the objective functions evaluated till this point.

**Step 4.4**

Calculate $J_{sw}$, i.e. the cost function value $(J)$ which is to be added with the swarm attractant cost $(J_{cc})$. So $(J_{sw})$ can be expressed by the following equation:

$$J_{sw}(i, j, k, ell) = J(i, j, k, ell) + J_{cc}\left(X_{gn}(j, k, ell), X(j, k, ell)\right)$$

**Step 4.5**

Let $J_{last} = J_{sw}(i, j, k, ell)$ save this value for finding a better cost via run.

**Step 4.6**

Tumble: Generate a random vector $\Delta(i)$ with each element $\Delta_n(i)$, $n = 1, 2, ..., m$, a random number in $[-1, 1]$.

**Step 4.7**

Move:

$$X^i(j+1, k, ell) = X^i(j, k, ell) + C(i)\frac{\Delta(i)}{\sqrt{\Delta^T(i)\Delta(i)}}$$

which results in a step of size $C(i)$ in the direction of the tumble for $i^{th}$ bacterium.

**Step 4.8**

Calculate $J(i, j+1, k, ell)$ and let $J_{sw}(i, j+1, k, ell) = J(i, j, k, ell) + J_{cc}\left(X^i(j+1, k, ell), P(j+1, k, ell)\right)$

**Step 4.9**

Swim:

    set m=0 (counter for swim length).

    While m<Ns (if have not climbed down too long ).

        m=m+1.

        If $J_{sw}(i, j+1, k, ell) < J_{last}$ (if doing better),

        Let $J_{last} = J_{sw}(i, j+1, k, ell)$ and

$$X^i(j+1,k,ell) = X^i(j,k,ell) + C(i)\frac{\Delta(i)}{\sqrt{\Delta^T(i)\Delta(i)}}$$

And use this $X^i(j+1,k,ell)$ to calculate the new $J_{sw}(i, j+1, k, ell)$.

    Else, set m=Ns.

    Go to next bacterium $(i+1)$ if $i \neq S$.

That is go to **step 4.2**; process the next bacterium.

**Step 5**

if $j < Nc$, go to **step 4.** In this case, continue chemotaxis since the life of the bacteria is not over.

**Step 6**

Reproduction:

    **Step 6.1**

    For the given $k$ and $ell$, and for each $i=1,2,...,S$, compute $J_{health}$ value.

    **Step 6.2**

    Sort the bacteria in order of descending $J_{health}$ value.

    **Step 6.3**

The Sr bacteria with the lowest $J_{health}$ values die and the remaining Sr bacteria with best values split (it is performed by the copies that are already made, now placed at the same location as their parents).

**Step 7**

If $k < N_{re}$, go to **step 3.** The next generation of the bacteria starts.

**Step 8**

Elimination-dispersal:

For $i = 1, 2, ..., S$, eliminate and disperse each bacterium with probability $P_{ed}$. To do this, if a bacterium is eliminated, simply disperse another one to a random location on the optimization domain. If $ell < N_{ed}$, go to **step 2**; otherwise, stop.

4. **Proposed technique**

The flowchart of fig-1 shows the proposed technique of segmenting the MR images. An MR image is taken and optimal multilevel thresholding is performed based on its histogram. Then the edges in the thresholded image are detected by comparing the intensity values. Eight connected neighbourhood pixels are considered and intensity value of the centre pixel is compared with each of the neighbouring pixels. Whenever we get a difference of nonzero value, one of the pixels is marked as edge pixel. The procedure is performed throughout the whole image to get the complete edge image. Then post processing is performed to get the final segmented image.

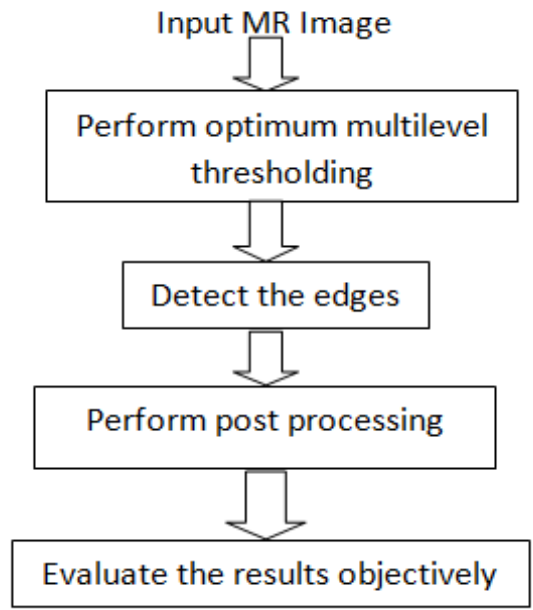

**Figure 1:** The proposed technique

*4.1 Determine the number of thresholds*

There are four contrast levels in brain MR Images (T1 and T2) corresponding to fat, white matter, grey matter and cerebrospinal fluid [14]. Histogram of the image should carefully be examined before determining the number of thresholds. We have experimented with various numbers of thresholds on a number of T1 and T2 weighted brain MR images. Usually, 3-4 level of thresholding in brain MR (T1 and T2) Images works well.

*4.2 Perform multilevel thresholding*

After selecting the suitable number of thresholds, the MR image is thresholded using bacterial foraging optimization algorithm by maximizing the objective function which in this case is *between class variance* as given by the equation -1 as discussed in section-2.

*4.3 Detect the edges*

Edges are detected by simply comparing the intensities in the thresholded image. To search for the edge pixels and mark them, we compute the intensity difference of the pixel $I(x, y)$ with the eight connected neighborhood in the thresholded image. Wherever we find that the intensity difference is nonzero, the greater intensity pixel is marked as edge (or the lower intensity pixel can also be marked as edge). It must be remembered that only one of them should be marked to map it with an edge. The intensity comparison is performed throughout the whole image to obtain the complete edge image.

*4.4 Post processing*

Post processing is done on the segmented image. The function '*bwareaopen* (*bw,p*)' in MATLAB is used to remove small objects (such as spurious and noise contents) in the segmented image. This function removes all connected components (objects) having less than *p* number of pixels.

*4.5 Objective evaluation*

The accuracy of the segmentation is measured using objective evaluation techniques such as "*figure of merit*", "*Rand Index*", and "*Variation of information*".

## 5. Results and discussion

Brain MR Image segmented results are presented in fig-2 to fig-9. The evaluation criteria *Pratt figure of merit, rand index* and *variation of information* are used to assess the performance of the proposed segmentation technique. A greater value of *figure of merit* specifies better performance of the edge detector. The Sobel and Canny edge is obtained using matlab toolbox. The figure of merit value obtained for BFO edge image is **0.8135** which is greater than the corresponding value for Canny and Sobel edge images as given in table1. The table2 gives the *Rand Index*, and *variation of information* values for BFO, Canny and Sobel edge images. Highest *Rand Index* and the lowest *Variation of Information* are achieved in the proposed BFO segmentation technique. A greater value of *Rand Index* and lesser value of *variation of information* indicate better performance of the segmentation method. Thus the parameter values obtained indicate better performance of the proposed technique than the traditional Canny and Sobel methods. Figure-9 shows the convergence of the BFO algorithm as the graph becomes progressively flat towards the right hand side.

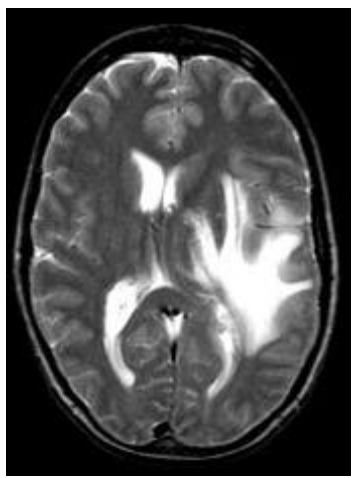 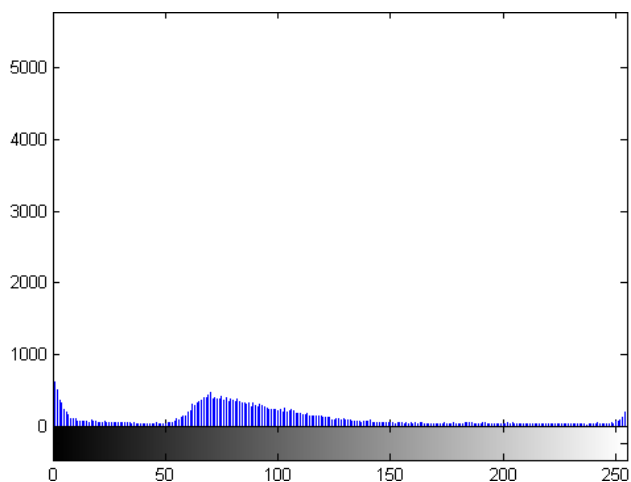

**Figure 2:** T2w brain MR image        **Figure 3:** Histogram of the image

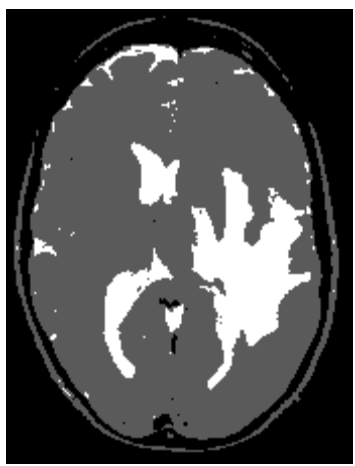 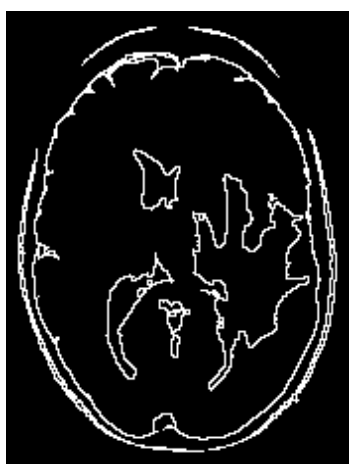 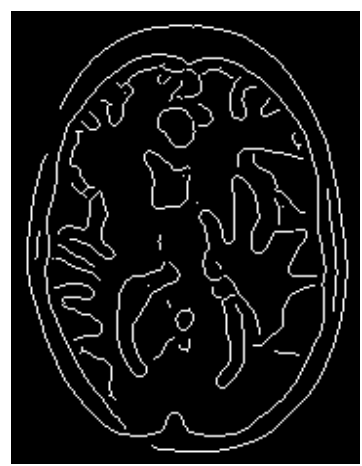

**Figure 4:** thresholded image    **Figure 5:** BFO segmented image    **Figure 6:** Canny edge image

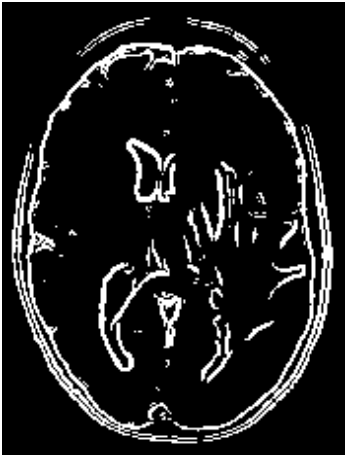 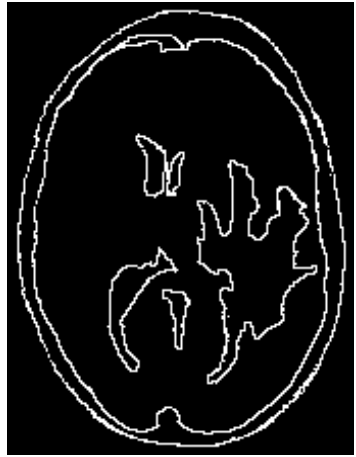 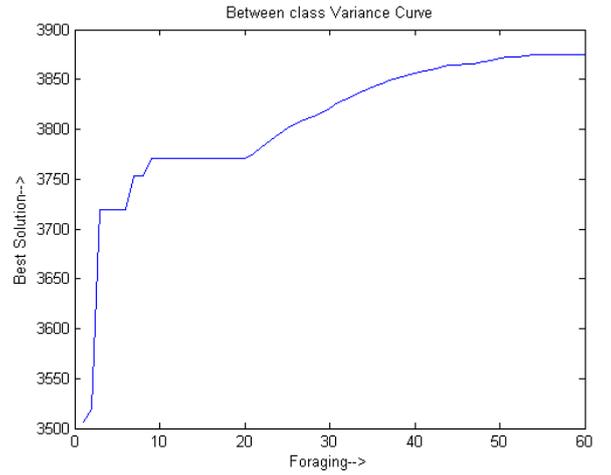

**Figure 7:** Sobel edge image       **Figure 8:** Manually segmented image       **Figure 9:** Convergence of the BFO algorithm

**Table 1:** Objective evaluation of the edge detectors

| Edge detector | Figure of merit |
|---|---|
| Canny | *0.4696* |
| Sobel | *0.3831* |
| Proposed | *0.8135* |

**Table 2:** Objective evaluation of the edge detectors

| Parameters | Sobel | Canny | Proposed |
|---|---|---|---|
| Rand Index (RI) | *0.8913* | *0.8988* | *0.9639* |
| Variation of information(VI) | *0.4647* | *0.4092* | *0.2102* |

## 6. Conclusions

This paper presents a multilevel thresholding based brain MR image segmentation technique for T1 and T2 weighted brain MR images. The visual quality of the segmented results and the values of the objective evaluation parameters demonstrate the superiority of the proposed technique over traditional methods such as canny and sobel. We have achieved a significantly higher *figure of merit* by the proposed technique than that of *canny* and *sobel* edge detection techniques. Moreover, the RI and VI parameters as obtained by the proposed method are also better than the *canny* and *sobel* edge detection techniques for T1 and T2 weighted brain MR images.